\documentclass[journal]{IEEEtran}
\ifCLASSINFOpdf
\else
\fi
\usepackage{amsmath,graphicx}
\usepackage{booktabs}
\usepackage{multicol}
\usepackage{multirow}
\usepackage{threeparttable}
\usepackage{newtxmath}
\usepackage{subcaption}
\begin{document}
%
\title{The style transformer with common knowledge optimization for image-text retrieval}
%
%
%

\author{Wenrui Li,
        Zhengyu Ma,
        Jinqiao Shi,
        and Xiaopeng Fan,~\IEEEmembership{~Senior Member,~IEEE}
\thanks{This work was supported in part by the National Key R\&D Program of China (2021YFF0900500), and the National Natural Science Foundation of China (NSFC) under grants U22B2035 and 62206141. (Corresponding author: Jinqiao Shi.)}
\thanks{Wenrui Li and Xiaopeng Fan are with the Department of Computer Science and Technology, Harbin Institute of Technology, Harbin 150001, China (e-mail: liwr@stu.hit.edu.cn; fxp@hit.edu.cn).}
\thanks{Zhengyu Ma is with the Peng Cheng Laboratory, Shenzhen 518055, China (e-mails: mazhy@pcl.ac.cn).}
\thanks{Jinqiao Shi is with the School of Cyberspace Security, Beijing University of Posts and Telecommunications, Beijing 100876, China (e-mails: shijinqiao@bupt.edu.cn).}}

%
%

\markboth{Journal of \LaTeX\ Class Files,~Vol.~14, No.~8, August~2015}%
{Shell \MakeLowercase{\textit{et al.}}: Bare Demo of IEEEtran.cls for IEEE Journals}
%



\maketitle

\begin{abstract}
Image-text retrieval which associates different modalities has drawn broad attention due to its excellent research value and broad real-world application. However, most of the existing methods haven't taken the high-level semantic relationships (“style embedding”) and common knowledge from multi-modalities into full consideration. To this end, we introduce a novel style transformer network with common knowledge optimization (CKSTN) for image-text retrieval. The main module is the common knowledge adaptor (CKA) with both the style embedding extractor (SEE) and the common knowledge optimization (CKO) modules. Specifically, the SEE uses the sequential update strategy to effectively connect the features of different stages in SEE. The CKO module is introduced to dynamically capture the latent concepts of common knowledge from different modalities. Besides, to get generalized temporal common knowledge, we propose a sequential update strategy to effectively integrate the features of different layers in SEE with previous common feature units. CKSTN demonstrates the superiorities of the state-of-the-art methods in image-text retrieval on MSCOCO and Flickr30K datasets. Moreover, CKSTN is constructed based on the lightweight transformer which is more convenient and practical for the application of real scenes, due to the better performance and lower parameters.  
\end{abstract}


%
\IEEEpeerreviewmaketitle

\section{Introduction}
\label{sec:intro}

\IEEEPARstart{I}mage-text retrieval \cite{sun1}--\cite{song1} refers to the process of retrieving relevant images and text from a database containing both modalities. It is an interdisciplinary field that combines computer vision \cite{intro4} and natural language processing \cite{BERT,transformer} and has been extensively studied. Most previous methods pay more attention to fine-grained alignment \cite{song2}--\cite{Litcsvt22} instead of the high-level semantic relationships. These methods not only lack the utilization of global information but also suffer from a large number of parameters. Besides, most of the current approaches \cite{intro1}--\cite{spl3} ignore the common knowledge among different modalities, which is significantly important for understanding the meaning of words in context and the feature representation ability of the network.

We proposed a novel style transformer network with common knowledge optimization (CKSTN) to tackle the above issues. To enhance the practicality of the model for real-world applications, we incorporate a lightweight feed-forward network instead the attention layer in transformer to reduce the model's parameters and computation costs. The style embedding extractor (SEE) is proposed to better utilize the high-level semantic relationships, which extract “style embedding” of feature maps row by row using the multi-layer perceptron (MLP). The concatenation and shuffle operations \cite{DeLighT} are utilized to ensure the “style embedding” contains more global information and shares different row features between feature maps. The style embedding extractor is a recurrent architecture to extract the multi-level features by fusing low-level and high-level semantic features effectively. The ``style embedding" not only contains the semantic concepts in high-level but also incorporates the pixel-level alignment relationships in low-level features. As for the high-level semantic information, we also employ common knowledge across modalities, which can effectively improve the feature representation as well. In the end, we propose the common knowledge optimization (CKO) module to dynamically update the latent concepts of common knowledge from the “style embedding” of different modalities, which can benefit the feature adjustment module. The sequential update strategy in CKO is proposed to dynamically adjust the common knowledge, which optimizes the feature representation among different modalities based on the style embedding of different items. We summarize the main contributions of this paper as follows:

\textbf{1)} We propose a novel style transformer network with common knowledge optimization (CKSTN), which utilized the common knowledge dynamically adjusted based on “style embeddings” to optimize the feature representations of two modalities. 

\textbf{2)} The SEE uses the concatenation and shuffle operation to share different row features across feature maps, and MLP to get compacted efficient representations. The CKA is proposed to capture the latent concepts of common knowledge across different modalities.

\textbf{3)} 
The sequential update strategy is introduced to integrate the features of different stages in SEE with previous common feature units, which helps to get generalized temporal common knowledge. 


\begin{figure*}
	\centering
	\includegraphics[scale=0.78]{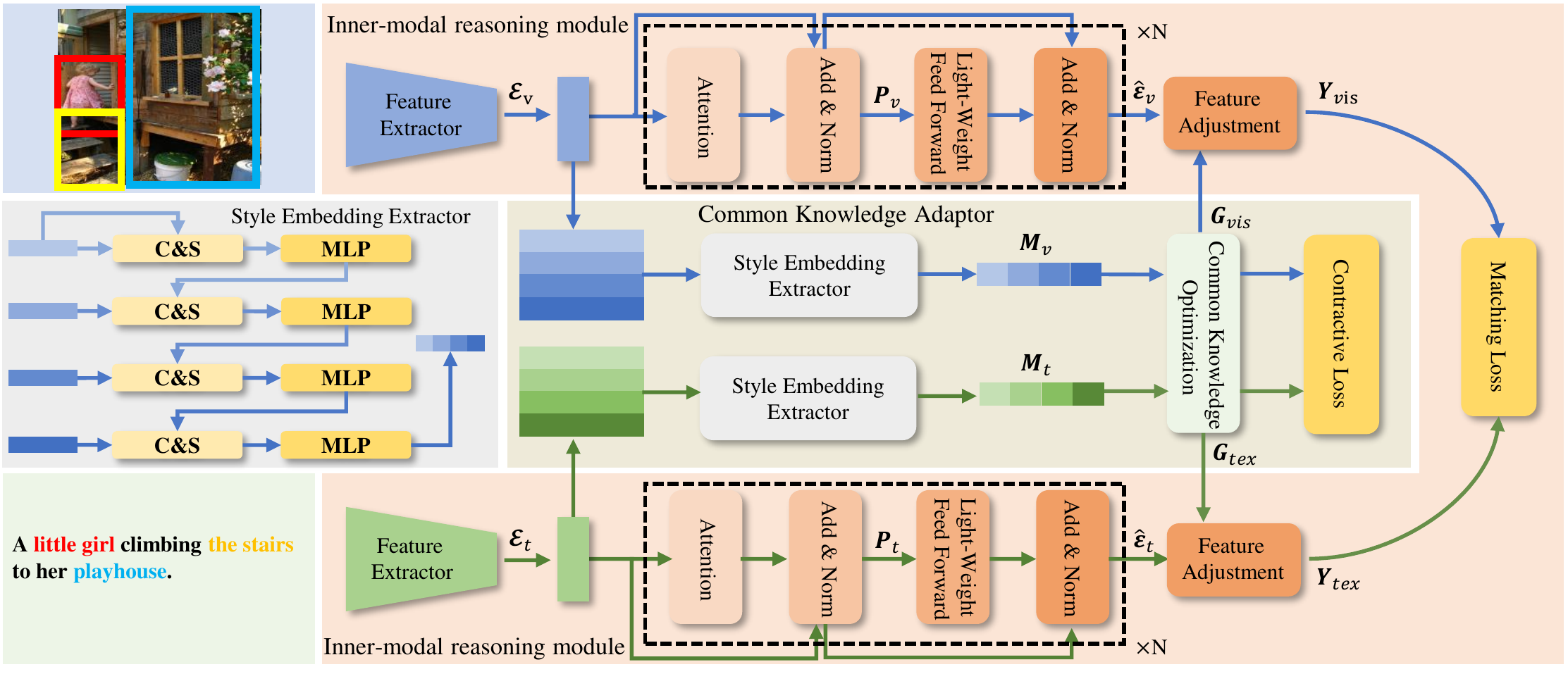}
	\caption{Architecture of the CKSTN. The “C\&S” represents concatenation and shuffle operation. The SEE could extract the high-level semantic relationships that represent global information effectively. The CKA could dynamically capture the common latent concepts of different modalities. Finally, the learnt common knowledge is utilized to further assist the formation of the item feature representations.}
	\label{fig:2}
\end{figure*}
\section{Approach}
\label{sec:format}
\subsection{Architecture}
\textbf{Feature extractor of image and text} RAVLN uses Faster-RCNN \cite{Faster-RCNN} to extract visual features and pre-trained BERT \cite{BERT} to extract textual features following \cite{LTAN} respectively. The information related to geometry bonding boxes is integrated into visual features extracted from images using a fully connected neural network. In pre-trained BERT, the standard transformer architecture could use the self-attention mechanism to encode the memory connections between words in sentences. Given the visual and textual features input $\boldsymbol{\mathcal{X}}_{c}\in \mathbb{R}^{n\times d_{o}}$, $n$ and $d_{o}$ are the sequence length and dimension of $\boldsymbol{\mathcal{X}}_{c}$, and $c\in (\boldsymbol{v},\boldsymbol{t})$. The outputs of the visual and textual feature extractor can be written as: $\boldsymbol{\mathcal{E}}_{v}=E_{vis}(\boldsymbol{\mathcal{X}}_{v})$ and $ \boldsymbol{\mathcal{E}}_{t}=E_{tex}(\boldsymbol{\mathcal{X}}_{t})$, where $\boldsymbol{\mathcal{E}}_{v} \in \mathbb{R}^{n\times d_{in}}$ and $\boldsymbol{\mathcal{E}}_{t} \in \mathbb{R}^{n\times d_{in}}$.

\textbf{Inner-modal reasoning module.} After the feature extractor, the inner-modal semantic relationships are further explored using a stack of lightweight transformers. A complete lightweight transformer layer in visual pipeline can be written as:
\begin{equation}
	\begin{aligned}
		\boldsymbol{\mathcal{E}}_{v}^{a} &= \frac{(\boldsymbol{\mathcal{E}}_{v}\boldsymbol{T}_{Q}\cdot (\boldsymbol{\mathcal{E}}_{v}\boldsymbol{T}_{K})^{T})}{\sqrt{d_{e}}}\boldsymbol{\mathcal{E}}_{v}\boldsymbol{T}_{V},
		\\\boldsymbol{P}_{v} &= \mathrm{Add\&Norm}(\boldsymbol{\mathcal{E}}_{v}^{a},\boldsymbol{\mathcal{E}}_{v}),
		\\\hat{\boldsymbol{\mathcal{E}}_{v}} &= \mathrm{Add\&Norm}(\mathrm{FFN}(\boldsymbol{P}_{v}),\boldsymbol{P}_{v}),
	\end{aligned}
\end{equation}
where $\boldsymbol{T}_{(\cdot)}\in \mathbb{R}^{d_{in}\times d_{e}}$ denote the learnable linear projections of query, key and value, $d_{e}$ represents the dimension of the common space, $\mathrm{Add\&Norm}(\cdot)$ contains a layer normalization operation and a residual connection, and $\mathrm{FFN}(\cdot)$ denotes the lightweight feed-forward network, which composed of two nonlinear layers with GLUE activation function.

\textbf{Common knowledge adapter.} Rational use of high-level semantic information can optimize feature representation. The SEE is proposed to encode the semantic relationships of high-level features (``style embedding"). The outputs of $i$-th layer of SEE is defined as:
\begin{equation}
	\begin{aligned}
		\boldsymbol{R}_{i} = \mathrm{C\&S}(\mathrm{CLIP}_{i}(\boldsymbol{\mathcal{E}}_{v}),\boldsymbol{M}_{i-1}), \boldsymbol{M}_{i} = \mathrm{MLP}(\boldsymbol{R}_{i}),
	\end{aligned}
\end{equation}
where $\mathrm{CLIP}_{i}(\cdot): \mathbb{R}^{n\times d_{e}} \rightarrow \mathbb{R}^{n\times (d_{e}/m)}$ represents the clip function to cut and return the $i$-th row of the feature map, and $m$ is the number of layers in SEE, The $\mathrm{C\&S}(\cdot,\cdot): (\mathbb{R}^{n\times d_{m}},\mathbb{R}^{n\times d_{m}}) \rightarrow \mathbb{R}^{n\times (2*d_{m})}, d_{m}=d_{e}/m$, denotes the concatenation and shuffle operation, which could contain more global information and share different row features between feature maps and $\mathrm{MLP}(\cdot): \mathbb{R}^{n\times (2*d_{m})} \rightarrow \mathbb{R}^{n\times d_{m}}$ is a multi-layer perceptron consists of 3-layer linear layers.

The common knowledge optimization (CKO) consists of a set of common feature units $\boldsymbol{S} = \{\boldsymbol{S}_{1},\boldsymbol{S}_{2},\ldots,\boldsymbol{S}_{k}\}$ which represents the common knowledge between two modalities. The illustration of CKO is shown in Fig. 2. For each visual and textual input $\boldsymbol{M}_{c}\in \mathbb{R}^{n\times d_{m}}$ and $c\in (\boldsymbol{v},\boldsymbol{t})$ from the last layer of SEE. The CKO could compute the importance of each common feature unit according to the input vectors and effectively combine them together, which can be written as:
\begin{equation}
		\boldsymbol{S}_{o} =\sum_{i=1}^{k} \mathrm{Softmax}(\boldsymbol{S}_{i}\boldsymbol{M}_{c})\boldsymbol{M}_{c},
\end{equation}
where $\mathrm{Softmax}(x) = 1/(1+e^{-x})$. The memory gate is proposed to selectively remain and exclude the information of fused features. The memory gate in visual pipeline is defined as follows:
\begin{equation}
    \boldsymbol{G}_{vis} = \mathrm{ReLU}(\boldsymbol{W}_{o}\boldsymbol{S}_{ovis}+b_{o}),
\end{equation}
where $\boldsymbol{W}_{o}\in \mathbb{R}^{n\times n}$ is learnable weight metric and $b_{o}$ is the bias item. After the memory gate, the style embeddings are integrated with the common knowledge reasonable.

In order to extract the effective common knowledge, we propose a sequential update strategy to connect the features of different stages in SEE with previous common feature units $\boldsymbol{S}_{t-1}\in \mathbb{R}^{n\times d_{m}}$. The common feature units are updated as follows:
\begin{equation}
    \boldsymbol{S}_{t} = (\boldsymbol{G}_{vis}\boldsymbol{G}_{tex})\boldsymbol{S}_{t-1} + (1-\boldsymbol{G}_{vis}\boldsymbol{G}_{tex})\boldsymbol{S}_{ovis}\boldsymbol{S}_{otex}.
\end{equation}
Finally, the features reasoned by lightweight transformers will be globally adjusted based on the style embeddings. The feature adjustment block in visual pipeline is defined as:
\begin{equation}
	\boldsymbol{Y}_{vis} = \mathrm{Softmax}(\boldsymbol{W}_{g}\boldsymbol{G}_{vis})\boldsymbol{\hat{\mathcal{E}}}_{v},
\end{equation}
where $\boldsymbol{W}_{g}\in \mathbb{R}^{n\times n}$ is learnable weight metric to fine-turning the outputs of the memory gate.

\begin{figure}
	\centering
	\includegraphics[scale=0.7]{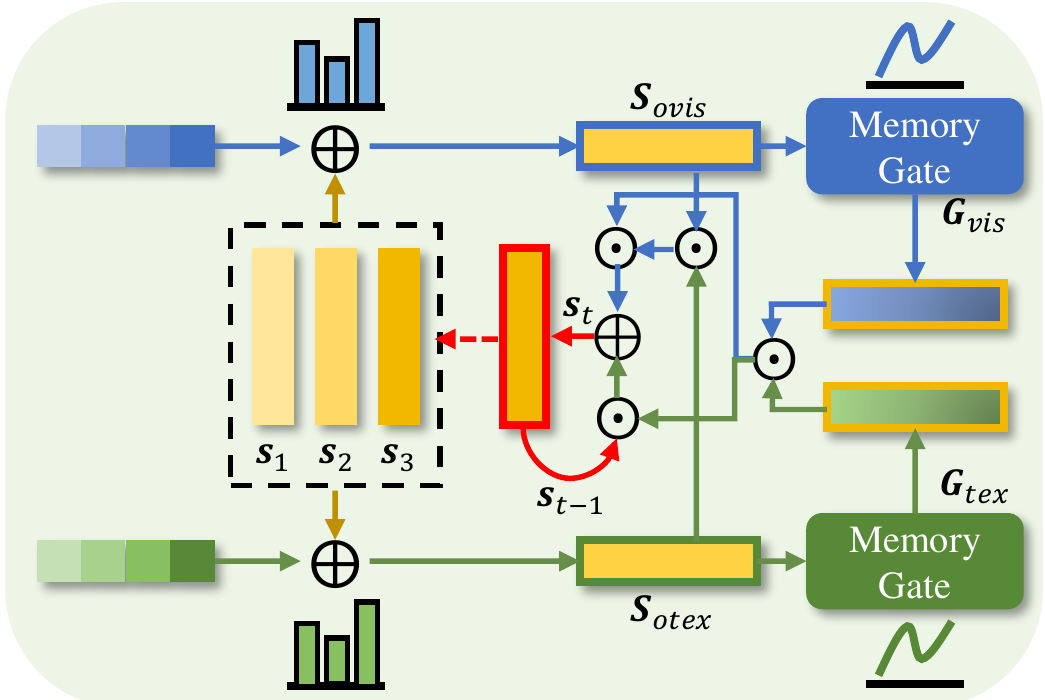}
	\caption{Architecture of the common knowledge optimization. The central red lines represent the sequential update strategy of $\boldsymbol{s}_{t}$.
Update details could be found in Eq. (3-5). The extracted $\boldsymbol{s}_{t}$ is fused with visual and textual style embedding to update and optimize the next input of visual and textual features.}
	\label{fig:2}
\end{figure}

\subsection{Training strategy}
We trained our model for 30 epochs on a single NVIDIA V100s GPU, the optimizer we choose Adam. We perform the bottom-up visual features extracted as proposed in \cite{intro4}, and mapped the features of two modalities into the dimension of embedding space which is 1024 to explore their relationships using a lightweight transformer and SEE module. To avoid overfitting, we utilized a warm-up strategy with a learning rate ranging from $1e^{-5}$ to $1e^{-4}$ for the first 10 epochs, which was then linearly decreased to $1e^{-5}$. We set the number of layers in the style embedding layers to 4, and for triplet ranking loss we used a hard negative margin of 0.2. To encourage the model to learn more accurate feature representations and increase the distance between mismatched pairs in the embedding space, the contrastive and matching loss is conpute based on the visual features and textual features.

\textbf{Contrastive loss.} A batch containing $N$ pairs of images and sentences, the intra-batch similarity between images and sentences can be defined as:
 \begin{equation}
\boldsymbol{\mathcal{L}}_{i2t}=-\frac{1}{N}\sum_{i=1}^{N}log\frac{exp((\boldsymbol{G}_{vis})_{i}^{T}(\boldsymbol{G}_{tex})_{i}/\tau )}{\sum_{N}^{j=1}exp((\boldsymbol{G}_{vis})_{i}^{T}(\boldsymbol{G}_{tex})_{i}/\tau )}.
\end{equation}
The definition of sentence-to-image similarity is shown as:
\begin{equation}
    \boldsymbol{\mathcal{L}}_{t2i}=-\frac{1}{N}\sum_{i=1}^{N}log\frac{exp((\boldsymbol{G}_{tex})_{i}^{T}(\boldsymbol{G}_{vis})_{i}/\tau )}{\sum_{N}^{j=1}exp((\boldsymbol{G}_{tex})_{i}^{T}(\boldsymbol{G}_{vis})_{j}/\tau )}.
\end{equation}
The contrastive loss is formulated as $ \boldsymbol{\mathcal{L}}_{con} = \boldsymbol{\mathcal{L}}_{i2t} + \boldsymbol{\mathcal{L}}_{t2i}$

\textbf{Matching loss.} We utilized a hinge-based triplet ranking loss \cite{BERT} to calculate the matching loss, which emphasizes on the hard negatives $l^{'}$ and $k^{'}$. The region-word similarity matrix $\boldsymbol{C} \in \mathbb{R}^{dk\times dl}$ is obtained by computing the cosine similarity between $\boldsymbol{Y}{vis}$ and $\boldsymbol{Y}{tex}$, where $dk$ represents the set of indices of region features from the $k$-th image and $dl$ represents the indices of the $l$-th sentence words. The matching loss is defined as:
\begin{equation}
	\mathcal{L}_{kl}=\max _{l^{'}}[\gamma +\boldsymbol{C}_{kl^{'}}-\boldsymbol{C}_{kl}]_{+}+\max _{k^{'}}[\gamma +\boldsymbol{C}_{k^{'}l}-\boldsymbol{C}_{kl}]_{+},
	\label{LOSS}
\end{equation}
where is the margin parameter that minimum distance between negative pairs and visual-textual embeddings of truly matched, and $ [s]_{+}\equiv \max(s,0)$. 

The total loss is formulated as $\mathcal{L}_{all} = \mathcal{L}_{con}+\mathcal{L}_{kl}$.

\begin{table*}

\renewcommand\arraystretch{1.2}
	\centering
	\begin{threeparttable}
		
		\caption{Experiment results on two benchmark dataset.}
		\label{TAB1}
		\setlength{\tabcolsep}{6pt}{
\begin{tabular}{cccccccccccccccc}
\toprule[1pt]
\multirow{3}{*}{Models} & \multicolumn{7}{c}{MS-COCO Dataset}                      & \multicolumn{7}{c}{Flickr30K Dataset}           &           \\ \cline{2-16} 
 &
  \multicolumn{3}{c}{Image Retrieval} &
  \multicolumn{3}{c}{Sentence Retrieval} &
  \multirow{2}{*}{\textit{Rsum}} &
  \multicolumn{3}{c}{Image Retrieval} &
  \multicolumn{3}{c}{Sentence Retrieval} &
  \multirow{2}{*}{\textit{Rsum}} &
  \multirow{2}{*}{\textit{Venue}} \\ \cline{2-7} \cline{9-14}
                        & R@1  & R@5  & R@10 & R@1  & R@5  & R@10          &       & R@1  & R@5  & R@10 & R@1  & R@5  & R@10 &       &           \\ \midrule[1pt]
DREN \cite{DERN}                    & 61.8 & 88.2 & 96.0 & 76.6 & 95.7 & 98.3          & 516.6 & 53.6 & 81.5 & 87.0 & 74.3 & 93.5 & 96.2 & 483.8 & TCSVT 22  \\
SSAMT \cite{SSAMT}                   & 62.7 & 89.6 & 95.3 & 78.2 & 95.6 & 98.0          & 519.4 & 54.8 & 81.5 & 88.0 & 75.4 & 92.6 & 96.4 & 488.7 & ICMR 22   \\
CGMN  \cite{CGMN}                     & 63.8 & 90.7 & 95.7 & 76.8 & 95.4 & 98.3          & 520.7 & 59.9 & 85.1 & 90.6 & 77.9 & 93.8 & 96.8 & 504.1 & TOMM 22   \\
GraDual \cite{GraDual}               & 63.7 & 90.8 & 95.6 & 76.8 & 95.9 & 98.3          & 521.1 & 57.7 & 83.1 & 90.5 & 76.1 & 93.7 & 96.7 & 497.8 & WACV 22   \\
MACM \cite{MACM}                    & 63.1 & 90.5 & 95.6 & 77.2 & 96.7 & 98.5          & 521.6 & 58.8 & 84.1 & 90.0 & 77.6 & 94.5 & 97.1 & 502.1 & IJCNN 22  \\
DIME \cite{DIME}                   & 63.0 & 90.5 & 96.2 & 77.9 & 95.9 & 98.3          & 521.8 & 60.1 & 85.5 & 91.8 & 77.4 & 95.0 & 97.4 & 507.2 & SIGIR 21  \\
TERAN \cite{TERAN}                  & 65.0 & 91.2 & 96.4 & 77.7 & 95.9 & 98.6          & 524.8 & 59.5 & 84.9 & 90.6 & 75.8 & 93.2 & 96.7 & 500.7 & TOMM 21   \\
LTAN \cite{LTAN}                   & 67.5 & 92.2 & 97.4 & 79.6 & 97.9 & \textbf{99.7} & 534.3 & 60.5 & 85.1 & 91.1 & 76.5 & 93.5 & 96.5 & 503.2 & ICASSP 22 \\ \midrule[1pt]
CKSTN &
  \textbf{69.1} &
  \textbf{92.3} &
  \textbf{98.8} &
  \textbf{81.1} &
  \textbf{98.1} &
  99.3 &
  \textbf{538.7} &
  \textbf{62.3} &
  \textbf{86.8} &
  \textbf{91.3} &
  \textbf{78.3} &
  \textbf{92.9} &
  \textbf{96.9} &
  \textbf{508.5} &
\\ \bottomrule [1pt] 
\end{tabular}}

	\end{threeparttable}
\end{table*}

\begin{table}
\small
	\centering
\renewcommand\arraystretch{1.2}
 \begin{threeparttable}
		
		\caption{The methods comparison on Flickr30K dataset.}
		\label{TAB1}
				\setlength{\tabcolsep}{2pt}{
\begin{tabular}{ccccccc}
\toprule[1pt]
Models                                  & R@1           & R@5           & R@10          & \textit{Rsum}           & \ \textit{\#params}      & \textit{Venue}      \\ \midrule[1pt]
ViLBERT \cite{ViLBERT} & 58.2          & 84.9          & \textbf{91.5} & 234.6          & 275M          & NeurIPS 19 \\
TERAN \cite{TERAN}     & 59.5          & 84.9          & 90.6          & 235.0          & 216M          & TOMM 21    \\
LTAN \cite{LTAN}       & 60.5          & 85.1          & 91.1          & 236.7          & 181M          & ICASSP 22  \\ \midrule[1pt]
CKSTN                                   & \textbf{62.3} & \textbf{86.8} & 91.3          & \textbf{240.4} & \textbf{175M} &            \\ 
\bottomrule[1pt]
\end{tabular}
}
\end{threeparttable}
\end{table}
\section{Experiment}
\subsection{Results comparison}
Our model is compared with the state-of-the-art baselines on Flick30K \cite{young2014image} and MSCOCO \cite{lin2014microsoft} datasets. To unify the criteria, we only evaluate basic models without ensemble. We utilize the recall at rank k (R@K) as the evaluation metric to measure the retrieval performance of the compared methods. The TERAN \cite{TERAN}, ViLBERT \cite{ViLBERT}, and LTAN \cite{LTAN} use the transformer to get the attention map. The DIME \cite{DIME} uses the “rectified identity cell” and “global-local guidance cell” to extract the general knowledge which helps the interaction of different modalities. However, the general knowledge extracted by DIME can’t adjust dynamically. Our model uses transformer to extract the attention relationships between different modalities and proposes the CKO module to optimize feature formation dynamically.

Table 1 shows that our CKSTN achieves superior performance, outperforming most of the baselines. For MSCOCO, CKSTN consistently outperforms the state-of-the-art method DERN with a relative improvement of 11.8\% and 5.9\% on R@1 for image retrieval and sentence retrieval, respectively. Moreover, our model improves Rsum by 22.1. For Flickr30K, our method outperforms the state-of-the-art model LTAN by 3.1\% and 2.4\% on R@1 for image retrieval and sentence retrieval, respectively. Table 2 compares our model with transformer-based models, indicating that our model achieves better performance with fewer parameters. While ViLBERT has 275M parameters, our model has only 175M parameters. Considering the practicality and efficiency requirements for real-world applications, our approach is more feasible and convenient for deployment.

\subsection{Ablation study}
\textbf{The effectiveness of the CKO, SEE and matching loss.} To demonstrate the effectiveness of our proposed CKO and SEE for image-text retrieval, we compare the model without the CKO and SEE on the left of Fig. 3, denote as ``W/o common knowledge optimization" and ``standard" with 0 number of layers in SEE. As for CKO, We find that “standard” obtains a significant improvement of 2.5\% and 3.1\% in R@1 for image and sentence retrieval respectively in 4 layers of SEE. CKO can capture the latent concepts of common between different modalities, which can better help the formation of feature representation. The model in 4 layers of SEE gets 5.2\% relative gains on Rsum compared with the model without SEE. It is worth mentioning that in the model without CKO, increasing the number of layers in SEE instead improves the model performance, which is inconsistent with the change of the model with CKO. With CKO, the global information of the item is extracted from fewer SEE layers with common knowledge for macro regulation. Without CKO, the more detailed style embedding can better assist the feature representation.

\indent\textbf{The effects of different number of layers in style embedding extractor.} We designed the experiments to investigate the effect of the number of layers in the SEE, by incrementally increasing the number of layers to 2, 4, 8, and 16. The CKSTN model without any modification is denoted as "standard" on the left of Fig. 3. Our findings suggest that an increase in the number of layers can improve retrieval accuracy, within a certain range, from 2 to 4 layers. However, when the number of layers is further increased to 8, the model's representation ability starts to decrease.

\indent\textbf{The effects of contrastive loss.} We fine-tune the final model by using the contrastive loss in the CKA to measure the difference between the style embeddings of visual features and textual features. As shown on the left of Fig. 3, the results without contrastive loss are slightly lower than the "standard" results, as expected.

\indent\textbf{Further analysis compared with LTAN.} LTAN is also one of the latest state-of-the-art papers using the lightweight transformer to compress the model parameters. On the right of Fig. 3, the structure of LTAN would suffer the ``cliff effect" when the dimension of the intermediate layer is reduced. However, adding our proposed CKO module, not only relieves the ``cliff effect", but also optimizes the feature representation with the same dimensions in the intermediate layer.

\begin{figure}
		\centering
		\includegraphics[scale=0.25]{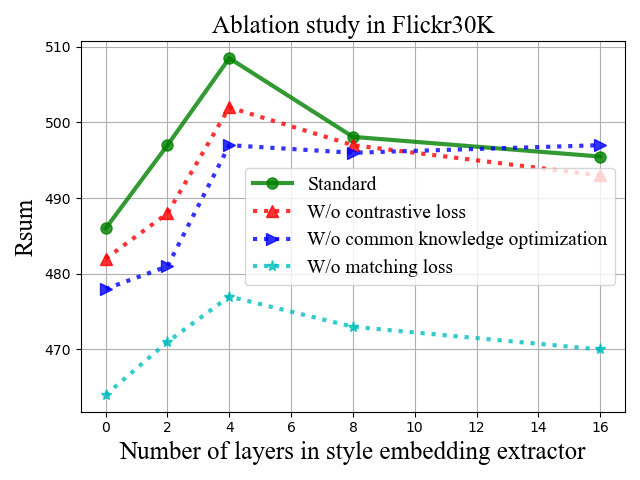}
		\includegraphics[scale=0.25]{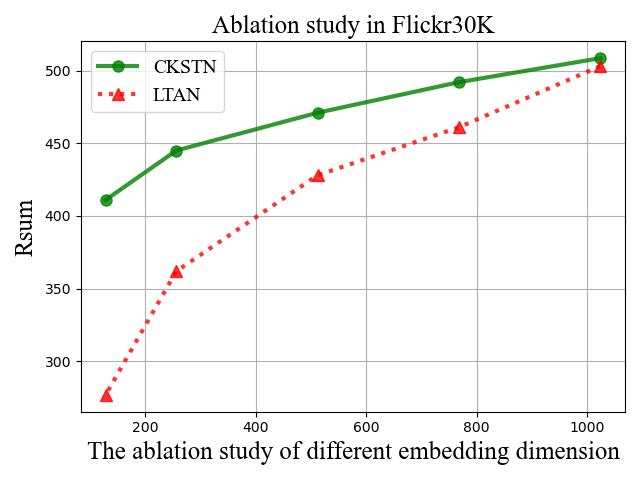}

 \caption{The ablation study on Flickr30K dataset.}
 \label{fig:2}
\end{figure}
\begin{figure}
	\centering
	\includegraphics[scale=0.45]{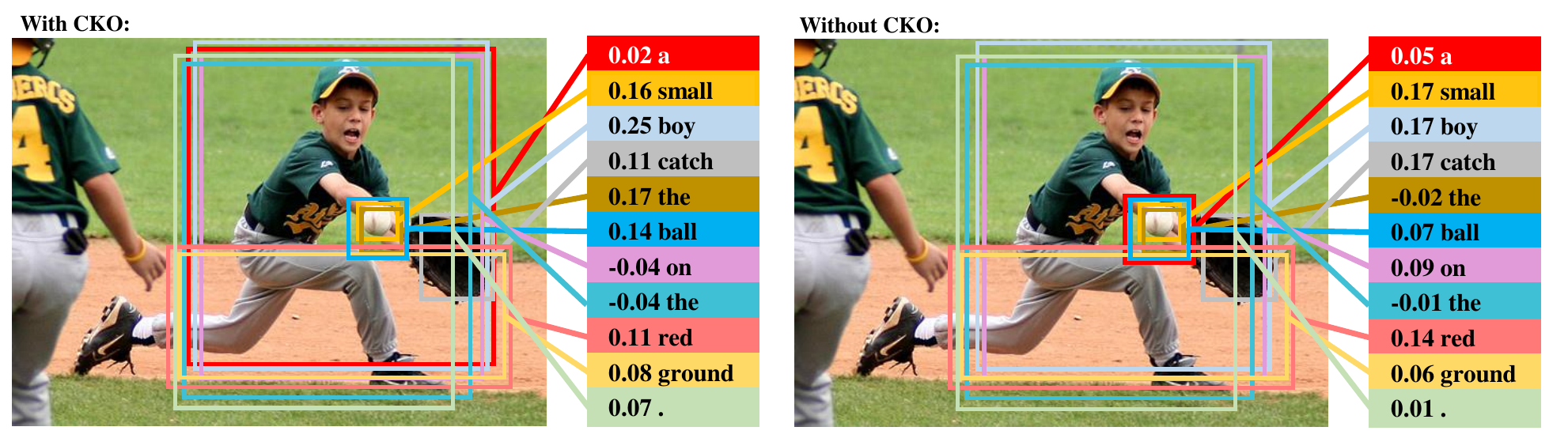}
	\caption{The visualization comparison of region-word matching.}
	\label{fig:2}
\end{figure}
\subsection{Visualization of Region-Word Matching}
In Fig. 4, we present the cosine similarity between every word and its corresponding image region with the highest association. The affinity between visual and textual features is measured by cosine similarity, just like in the training phase. Without the CKO, the similarity of alignment relationships of instance objects (e.g., boy, ball) will drop. See the red area in Fig. 4. The article “a” modifies “boy”. However, without the common knowledge optimization module, the article “a” makes mistakes of modifying “ball”. This also indicates that common knowledge contains global information to optimize the feature representation.

\section{conclusion}
The proposed style embedding module employs concatenation and shuffle operations to share distinct row features across feature maps. To capture the latent concepts of different modalities and dynamically adjust them based on the style embeddings, we propose common knowledge optimization operation. The experimental results validate the effectiveness of CKSTN, which outperforms previous state-of-the-art methods on two benchmark datasets while having fewer parameters than transformer-based methods.
\label{sec:refs}

%








\end{document}